\documentclass{article}

\usepackage{PRIMEarxiv}

\usepackage[utf8]{inputenc} % allow utf-8 input
\usepackage[T1]{fontenc}    % use 8-bit T1 fonts
\usepackage{hyperref}       % hyperlinks
\usepackage{url}            % simple URL typesetting
\usepackage{booktabs}       % professional-quality tables
\usepackage{amsfonts}       % blackboard math symbols
\usepackage{nicefrac}       % compact symbols for 1/2, etc.
\usepackage{microtype}      % microtypography
\usepackage{multirow}
\usepackage{amsmath}
\usepackage[title]{appendix}
\usepackage{fancyhdr}       % header
\usepackage{graphicx}       % graphics
\graphicspath{{media/}}     % organize your images and other figures under media/ folder

\title{SST-GCN: The Sequential based Spatio-Temporal Graph Convolutional networks for Minute-level and Road-level Traffic Accident Risk Prediction}

\author{
  Tae-wook Kim \\
  Division of Software \\
  Yonsei University \\
  Wonju, South Korea\\
  \texttt{xodnr8436@yonsei.ac.kr} \\
    \And
  Han-jin Lee \\
  Department of Computer Science \\
  Yonsei University \\
  Wonju, South Korea\\
  \texttt{han-.-jin@yonsei.ac.kr} \\
      \And
  Hyeon-Jin Jung \\
  Department of Computer Science \\
  Yonsei University \\
  Wonju, South Korea\\
  \texttt{wjdehdtod@tonsei.ac.kr} \\
      \And
  Ji-Woong Yang \\
  Artificial Intelligence Semiconductor \\
  Hanyang University \\
  Seoul, South Korea\\
  \texttt{jiwoong0412@hanyang.ac.kr} \\
      \And
  Ellen J. Hong\\
  Division of Software \\
  Yonsei University \\
  Wonju, South Korea\\
  \texttt{ellenhong@yonsei.ac.kr} \\
}

\begin{document}
\maketitle

\begin{abstract}
Traffic accidents are recognized as a major social issue worldwide, causing numerous injuries and significant costs annually. Consequently, methods for predicting and preventing traffic accidents have been researched for many years. With advancements in the field of artificial intelligence, various studies have applied Machine Learning and Deep Learning techniques to traffic accident prediction. Modern traffic conditions change rapidly by the minute, and these changes vary significantly across different roads. In other words, the risk of traffic accidents changes minute by minute in various patterns for each road. Therefore, it is desirable to predict traffic accident risk at the Minute-Level and Road-Level. However, because roads have close and complex relationships with adjacent roads, research on predicting traffic accidents at the Minute-Level and Road-Level is challenging. Thus, it is essential to build a model that can reflect the spatial and temporal characteristics of roads for traffic accident prediction. Consequently, recent attempts have been made to use Graph Convolutional Networks to capture the spatial characteristics of roads and Recurrent Neural Networks to capture their temporal characteristics for predicting traffic accident risk. This paper proposes the Sequential based Spatio-Temporal Graph Convolutional Networks (SST-GCN), which combines GCN and LSTM, to predict traffic accidents at the Minute-Level and Road-Level using a road dataset constructed in Seoul, the capital of South Korea. Experiments have demonstrated that SST-GCN outperforms other state-of-the-art models in Minute-Level predictions.
\end{abstract}

% keywords can be removed
\keywords{
         Traffic accident prediction
    \and Graph convolutional network 
    \and Intelligent transportation system
    \and Minute-level
    \and Road-level
}

\section{Introduction}
Traffic accidents are recognized as a significant social issue worldwide, causing both human casualties and property damage. Therefore, preventing traffic accidents to reduce social costs is a matter of great importance. Traffic conditions and accident risks change rapidly by the minute, and these changes vary significantly across different roads. Hence, it is necessary to predict the risk of traffic accidents at the Minute-Level and Road-Level to efficiently prevent and respond to such incidents. However, due to the complex and extensive nature of road conditions, predicting risk at a minute and road level is a challenging task.

In the past, studies have compared the performance of traffic accident prediction using Machine Learning models such as Decision Tree, Random Forest\cite{1}\cite{2}\cite{3}, and XGBoost\cite{4}. With the advent of Deep Learning technologies, research has been conducted using models based on Deep Neural Networks, Convolutional Networks, and Recurrent Neural Networks for traffic-related predictions. For Deep Neural Network-based models, a traffic accident risk prediction model using Fully Connected Layers\cite{5} was employed. Convolutional Neural Network-based research included models like AutoEncoder\cite{6} and models utilizing ResNet\cite{7}\cite{8}. For Recurrent Neural Networks, a Real-Time Traffic Accident Risk Prediction Model using LSTM layers\cite{9} was developed. Such research has continuously aimed to improve the accuracy of road traffic accident prediction using Deep Learning models. However, the models mentioned above did not consider the relationship between arbitrary roads and their adjacent roads.

To address these issues, it is important to construct and preprocess road information data and select models that are suitable for the characteristics of the roads. Since the node-link structure of roads is similar to the node-edge structure of a graph, roads can be effectively represented as graphs. Therefore, it is desirable to represent the spatial characteristics of roads using a graph and to utilize a Graph Convolutional Network (GCN)\cite{10} to effectively process the graph.

The Graph Convolutional Network (GCN) performs convolutional computations on graph data, applying the key feature of convolutional operations—weight sharing—to graph data. Ultimately, GCN is a specialized technique for modeling interactions between nodes and edges in a graph, enabling the learning of complex patterns. Additionally, since past conditions of the roads also influence traffic accidents, it is crucial to consider the temporal characteristics of roads. To effectively process sequential data with temporal characteristics, Recurrent Neural Network (RNN) models are commonly used, with significant research focused on Long Short-Term Memory (LSTM)\cite{11}. LSTM is characterized by its ability to remember past information and reflect it in current information, enabling it to learn patterns in sequential data. Specifically, LSTM is effective in capturing temporal dependencies based on previous traffic conditions, weather conditions, and other data, making it a suitable model for predicting traffic accident risk.

Therefore, to effectively predict traffic accident risk, it is necessary to have a model that can reflect both the spatial and temporal characteristics of roads. This paper proposes the Sequential based Spatio-Temporal Graph Convolutional Networks (SST-GCN), which combines GCN to capture the spatial characteristics of each road at the Minute Level and Road Level, and LSTM to reflect the temporal characteristics of the roads.

In addition to selecting an appropriate model for predicting traffic accident risk, it is crucial to collect and preprocess the dataset to enable more effective model training. Among the vast amounts of data representing road information, some data significantly influence traffic accident risk prediction, while others do not. Using all available data for prediction could degrade the model's performance compared to using only relevant data, leading to difficulties in prediction. Therefore, constructing a dataset based solely on information that impacts traffic accident risk is essential.

Furthermore, most studies only consider situational information such as average speed or traffic volume, neglecting environmental factors (e.g., solar azimuth and altitude, weather conditions). This paper experimentally demonstrates that environmental factors, including solar azimuth, altitude, and weather, also influence traffic accidents. 

Finally, during the preprocessing of the collected data to represent roads as graphs for Road-Level prediction, this paper proposes representing roads as nodes and intersections as edges. The study experimentally proves that expressing the graph's adjacency matrix as a normalized Laplacian matrix is more effective than representing the weights between roads as 0 or 1 based on adjacency. This approach better captures the relationships between roads, enhancing the accuracy of the traffic accident risk prediction.

To summarize, our main contributions are:

\begin{itemize}
    \item We introduce SST-GCN - The model that can predict the risk of traffic accidents at both the road-level and minute-level by incorporating the temporal and spatial conditions of the road, based on GCN and LSTM.

    \item We have experimentally shown that environmental factors such as solar azimuth angle, elevation, and weather have a positive impact on improving the performance of our model.

    \item We have experimentally demonstrated that representing roads as a graph structure using the normalized Laplacian matrix is more effective than using a regular adjacency matrix.
\end{itemize}

\section{Related Works}
\label{sec:headings}

\subsection{RNN-based Models}
A method for real-time traffic accident risk prediction was proposed by \cite{12}, introducing the long short-term memory convolutional neural network (LSTM-CNN) model. The LSTM-CNN model was trained using various features such as traffic flow characteristics, signal cycles, and weather conditions. Specifically, the LSTM effectively captured long-term relevant features, while the CNN captured features that were not time-dependent. To address the imbalance between accident and non-accident data, the Train Dataset was resampled using the synthetic minority over-sampling technique (SMOTE). Five commonly used models—XGBoost, Bayesian Logistic Regression, CNN, LSTM—were compared to the LSTM-CNN model in terms of performance. The LSTM-CNN model demonstrated superior performance in metrics such as Area Under the Curve (AUC), sensitivity, and false alarm rate. Notably, it was found that the LSTM-CNN model performed better when combined in a parallel structure rather than a sequential one.
In another study, \cite{13} proposed a method for predicting traffic accidents using Attention Techniques. The model, named 'regional Traffic Accident risk that utilizes a Spatial-Temporal Attention Network (TA-STAN),' was designed to learn the traffic impact of Local and Global Regions separately. To improve model accuracy, numerous external environmental factors were selected as features, and the real traffic accident dataset from New York City was used. Six models were employed for performance comparison, using evaluation metrics such as MSE, RMSE, and MAE. The TA-STAN model outperformed the comparative models across all three evaluation metrics.
However, the aforementioned RNN-based models have the limitation of only reflecting the temporal characteristics of roads, without incorporating the graph-based characteristics of road networks.

\subsection{GCN-based Models}
To predict accidents by considering spatial, temporal, and external factors, \cite{14} proposed the Deep Spatio-Temporal Graph Convolutional Network (DSTGCN) model. This model is composed of three main parts: the first part learns spatial information through graph convolution operations; the second part uses basic Convolutional Layers and Graph Convolutional Layers to learn features that consider both spatial and temporal information; the third part uses an Embedding Layer to learn external information. The model was trained and evaluated using a dataset constructed from accident data, urban traffic speed, road networks, weather information, and Points of Interest (POI) data. The DSTGCN model outperformed classical and state-of-the-art methods. However, while GCN-based models effectively incorporate the graph-based characteristics of roads, they have limitations in utilizing RNN-based models that are suitable for analyzing the temporal characteristics of roads.

\subsection{GCN+RNN-based Models}
To predict traffic risk using inputs such as real-time accident risk, real-time traffic volume, and real-time weather data for a local region, \cite{15} proposed the space gated memory network (SGMN) model. This model takes data from the past 8 hours as input to predict traffic accident risk for the next hour. The input data, which includes past accident risk and traffic volume, is structured in the form of a graph. The SGMN model is constructed using Graph Convolutional Layers (GCN) and Gated Recurrent Units (GRU). Experimental results demonstrated that the SGMN model outperformed comparative models including RNN, LSTM, GRU, Convolution, and Hereto-ConvLSTM models.
To predict traffic accidents by reflecting the spatiotemporal characteristics of roads at the road-level and minute-level, \cite{16} proposed the Multi-Attention Dynamic Graph Convolution Network (MADGCN). This model uses a graph convolution network combined with multi-attention techniques. The paper introduced a loss function designed to increase sensitivity to accident data by increasing the loss for misclassified accident data. Compared to state-of-the-art methods, the MADGCN model showed superior performance.
To enhance traffic accident prediction performance, \cite{17} adopted a multi-view graph neural network and proposed MG-TAR, an end-to-end framework for effectively learning the correlations of multiple graphs for accident risk prediction. Using driving log data and statistics on risky driving, the study conducted a spatiotemporal correlation analysis between risky driving and actual accidents. Additionally, besides the traditional adjacency matrix, other adjacency matrices were used to represent various dependencies between regions.

\section{Data Preparation}
\subsection{Data Collection}
In this study, we construct the SST-GCN using 29 types of features classified into 5 categories, collected from Seoul, a traffic hub in South Korea, where approximately 23,000 roads and around 16,000 traffic accidents occur monthly. The details of the collected data are as follows:

\begin{table}
 \caption{Key Features in the dataset}
  \centering
  \resizebox{\textwidth}{!}{
  \begin{tabular}{l l l l}
    \toprule
    Type     & Notation        & Range of Value    & Description \\
    \midrule
    \multirow{5}{*}{Traffic Accident}   
    & $x^{day\_of\_the\_week}$  
    & One hot vector, Size=7        
    & day of the week\\    
    
    & $x^{time}$
    & $[0, 1439] \in \mathbb{Z}$
    & accident Occurrence Time (Minute-Level) \\
    
    & $x^{season}$
    & One hot vector, Size=4
    & season \\
    
    & $x^{sun\_altitude}$
    & $[0, 76.122] \in \mathbb{R}$
    & sun Altitude \\
    
    & $x^{sun\_diff}$
    & $[0, 180] \in \mathbb{R}$
    & the angle between the sun and the road \\

    \midrule
    \multirow{3}{*}{Road Information}
    & $x^{lanes}$
    & $[1, 7] \in \mathbb{Z}$
    & number of lanes \\
    
    & $x^{speed\_limit}$
    & $[10, 110] \in \mathbb{N}$
    & speed limit of a road \\
    
    & $x^{length}$
    & $[12.6924, 9101.8543] \in \mathbb{R}$
    & the length of a road \\

    \midrule
    \multirow{2}{*}{Road Safety Facility}
    & $x^{bump}$
    & Flag, 0 or 1
    & presence or absence of a speed bump on a road \\
    
    & $x^{camera}$
    & Flag, 0 or 1
    & presence or absence of cctv on a road \\

    \midrule
    \multirow{10}{*}{Point of Interest}
    & $x^{golf}$
    & $[0, 10] \in \mathbb{Z}$
    & number of golf practice facility \\
    
    & $x^{sales}$
    & $[0, 10] \in \mathbb{Z}$
    & number of door to door sales \\
    
    & $x^{gym}$
    & $[0, 10] \in \mathbb{Z}$
    & number of fitness center \\
    
    & $x^{mail}$
    & $[0, 10] \in \mathbb{Z}$
    & number of mail order business \\
    
    & $x^{food}$
    & $[0, 10] \in \mathbb{Z}$
    & number of general restaurant \\
    
    & $x^{bakery}$
    & $[0, 10] \in \mathbb{Z}$
    & number of bakery \\
    
    & $x^{food\_center}$
    & $[0, 10] \in \mathbb{Z}$
    & number of group catering facility \\
    
    & $x^{restaurant}$
    & $[0, 10] \in \mathbb{Z}$
    & number of rest area food service \\
    
    & $x^{pharm}$
    & $[0, 10] \in \mathbb{Z}$
    & number of pharmacy \\
    
    & $x^{hospital}$
    & $[0, 10] \in \mathbb{Z}$
    & number of hospital \\

    \midrule
    Traffic speed
    & $x^{traffic\_speed}$
    & $[0, 110]$
    & the average speed of a road for five minutes \\

    \midrule
    \multirow{8}{*}{Weather}
    & $x^{rain}$
    & $[0, 64.7] \in \mathbb{R}$
    & amount of rainfall \\
    
    & $x^{temperature}$
    & $[-18.5, 36.3] \in \mathbb{R}$
    & temperature \\
    
    & $x^{humidity}$
    & $[11, 100] \in \mathbb{N}$
    & humidity \\
    
    & $x^{visibility}$
    & $[33, 5000] \in \mathbb{N}$
    & visibility \\
    
    & $x^{dew\_point}$
    & $[-27, 26.4] \in \mathbb{R}$
    & dew point \\
    
    & $x^{cloud}$
    & $[0, 10] \in \mathbb{Z}$
    & cloud \\
    
    & $x^{vaper\_press}$
    & $[0.7, 34.4] \in \mathbb{R}$
    & vapor pressure \\
    
    & $x^{ground\_temp}$
    & $[-12.7, 58.7] \in \mathbb{R}$
    & ground temperature \\

    \bottomrule
  \end{tabular}
  }
\end{table}

\paragraph{Traffic Accident Data}
was constructed using traffic accident data provided by the National Traffic Information Center (ITS)[18]. In this study, we used traffic accidents that occurred in the Seoul area from January 1, 2020, to December 31, 2021. For each traffic accident, the data includes the day of the week, time, season, and solar altitude at the time of the accident, as well as the angle between the road where the accident occurred and the solar azimuth.

\paragraph{Road Information Data}
was constructed using standard node-link data provided by the National Traffic Information Center (ITS)\cite{18}. This data includes general attributes of roads, which are considered to potentially impact traffic accidents, such as the number of lanes, speed limits, and road lengths.

\paragraph{Road Safety Facility Data}
was constructed using public data provided by the Ministry of the Interior and Safety\cite{19}. The Road Safety Facility data includes information on speed bumps and CCTVs, with both types of information assigned a value of 0 or 1 based on their presence or absence on a given road.

\paragraph{Point of Interest(POI) Data}
was collected based on industry group inquiry data provided by the Ministry of the Interior and Safety\cite{19}. This data consists of the 10 features most closely related to accidents, identified through frequency analysis of accidents and POIs. The number of specific POIs on a given road is limited to a maximum of 10.

\paragraph{Traffic Speed Data}
was constructed using traffic flow information data provided by the National Traffic Information Center (ITS)\cite{18}. This traffic flow information consists of the average speed of roads in 5-minute intervals. Therefore, this study utilized the 5-minute interval traffic speeds of roads in the Seoul area from January 1, 2020, to December 31, 2021.

\paragraph{Weather Data}
The Weather Data was constructed based on Automated Synoptic Observing System (ASOS) data provided by the Korea Meteorological Administration (KMA)\cite{20}. After performing a correlation analysis, only the data with a high correlation coefficient with traffic accidents were extracted. The resulting dataset includes precipitation, temperature, humidity, visibility, dew point, cloud cover, vapor pressure, and surface temperature.

Details on the correlation analysis, preprocessing process, and dataset construction can be found in our other paper \cite{dataset-paper}. The specifics of the constructed data are shown in Table 1.

\subsection{Construction of Traffic Accident Graph Dataset}
\begin{figure}
    \centering
    \includegraphics[width=\textwidth]{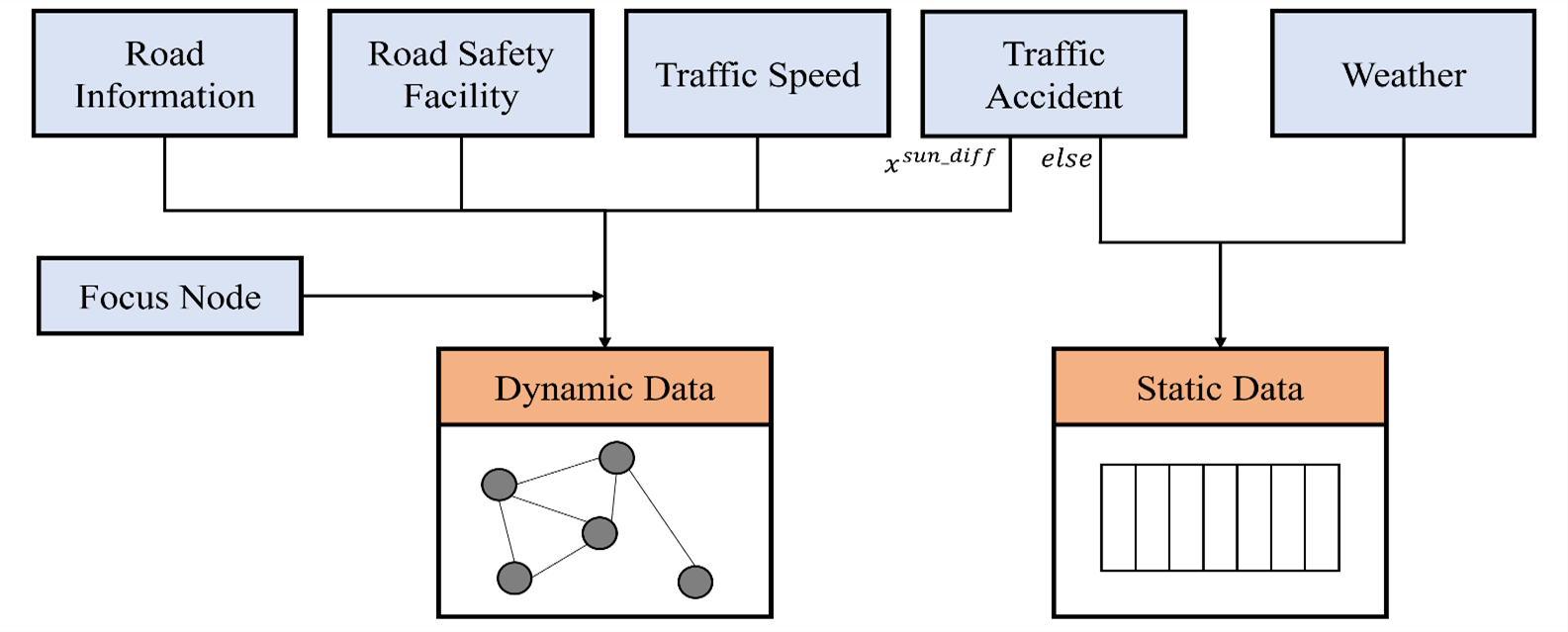}
    \caption{The Framework of the Dataset Construction}
    \label{fig:The Framework of the Dataset Construction}
\end{figure}
In this chapter, our objective is to construct a dataset \( D \) consisting of samples that include spatio-temporal features of multiple roads and the occurrence of accidents. For accident data, each data point includes the time the accident occurred and roads within a K-HOP distance from the accident site. For non-accident data, the dataset is constructed in the same manner, centering on random roads and random times. The determination of the K-HOP value, the number of sequences, and the time interval between sequences in the dataset is detailed in Chapter 4.4. Additionally, the ratio of accident to non-accident data in dataset \( D \) is set to 1:1.

The data consists of Graph Data, Static Data, and accident labels. Graph Data is composed of nodes extracted according to the K-HOP value, with each node having dynamic features that vary by road. Static Data contains static features that do not change within a local area and can be shared across all roads.

In Equation (1), \( X \) represents a single data point, where \( G_t \) and \( S_t \) denote the Graph Data and Static Data at time \( t \), respectively. \( y_{t+k} \) indicates whether the data is accident or non-accident at time \( t+k \). Here, \( n \) represents the number of sequences included in \( X \), and \( k \) denotes the minute-level time interval between sequences.

\begin{equation}
    X = ([G_{t-(n-1)k}, ..., G_t], [S_{t-(n-1)k, ..., S_t}], y_{t+1}) \in D
\end{equation}

Graph Data \( G_{(t-k)} \) consists of nodes \( V_{(t-k)} \), which represent dynamic features that change for each road at \( k \) minutes before the accident occurs, and the edges connecting these nodes with weights that form the Graph Normalized Laplacian matrix \( L \). This is expressed in Equation (2).

\begin{equation}
    G_{t-k} = (V_{t-k}, L)
\end{equation}

\( V_{(t-k)} \) is composed of nodes extracted based on the K-Hop criterion from a specific node. The \( i \)-th node in \( V_{(t-k)} \) contains the dynamic features \( v_{(t-k,i)} \) associated with the road linked to that node. Each node includes dynamic features from the Traffic Accident Data, such as the angle between the solar azimuth and the road, Road Safety Facility Data, Road Information Data, Point of Interest Data, Traffic Speed Data, and whether the node was the focal point for extracting the graph using the K-Hop criterion, represented by \( x_{(t-k,i)}^{focus} \). This is expressed in Equation (3).

\begin{equation}
    v_{t-k, i} = [
                  x^{sun\_diff}_{t-k,i},
                  x^{Road\_Information}_{t-k,i},
                  x^{Road\_Safety\_Facility}_{t-k,i},
                  x^{POI}_{i},
                  x^{traffic\_speed}_{t-k,i},
                  x^{focus}_{t-k,i}
                  ] \in V_{t-k}
\end{equation}

The Graph Normalized Laplacian matrix \( L \) is a matrix that applies a Normalized Laplacian Filter to the Distance Matrix \( A_{distance} \). The Distance Matrix \( A_{distance} \) has weights based on the distance from the road and includes the identity matrix \( I \) to consider the node's own features during GCN operations. This is expressed in Equation (4).

\begin{equation}
    L = I - D^{-1/2}A_{distance}D^{-1/2}
\end{equation}

Static Data \( S_{(t-k)} \) consists of the static features of the road at \( k \) minutes before the accident occurs. The features from the Traffic Accident Data, except for the angle difference between the sun and the road (\( x^{(sun\_diff)} \)), have the same values within a local region and can thus be classified as Static Data. Therefore, Static Data includes the four features from the Traffic Accident Data, excluding the solar azimuth angle difference, and the Weather Data. This is expressed in Equation (5), where \( k \) represents the minute-level time interval between sequences. The overall framework for constructing the dataset is shown in Figure 1.

\begin{equation}
    S_{t-k} = [
        x^{day_of_the_week}_{t-k},
        x^{time}_{t-k},
        x^{season}_{t-k},
        x^{sun_altitude}_{t-k},
        x^{Weather}_{t-k}
    ]
\end{equation}

\section{SST-GCN}

\begin{figure}
    \centering
    \includegraphics[width=\textwidth]{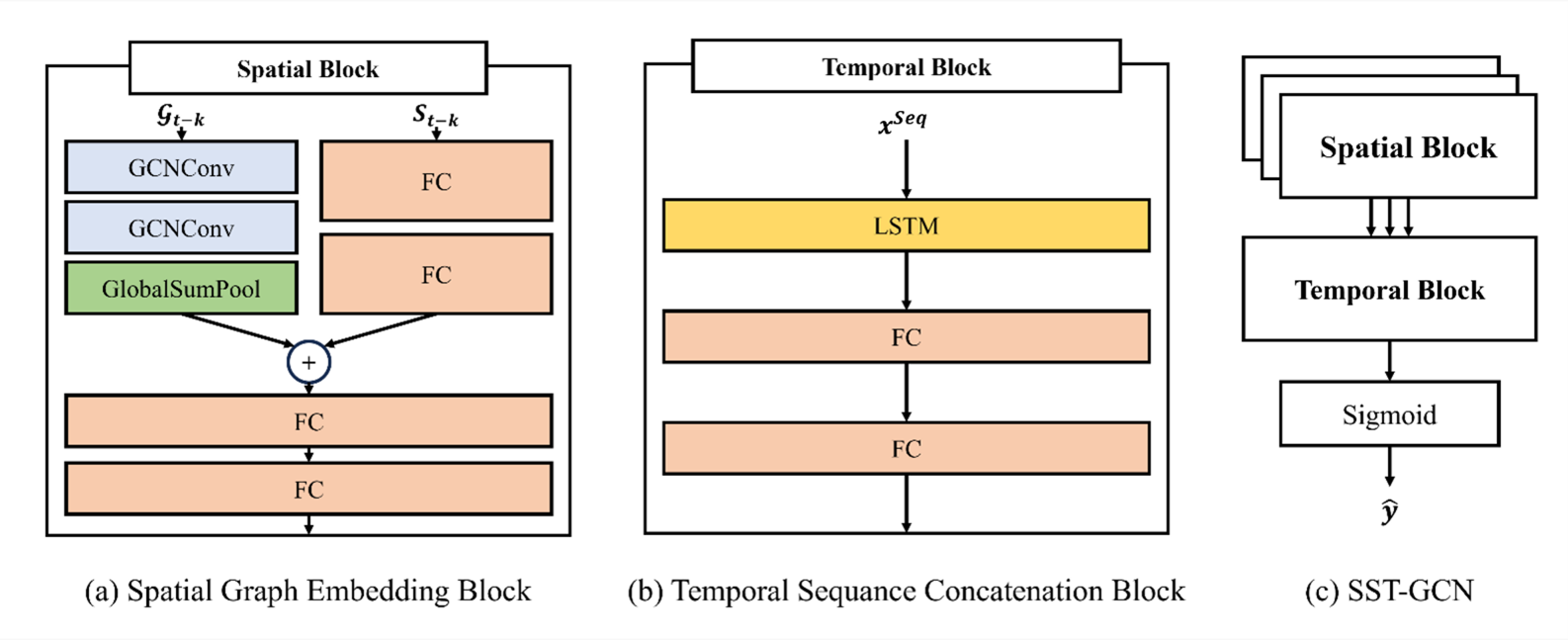}
    \caption{The Architecture of SST-GCN Model}
    \label{The Architecture of SST-GCN Model}
\end{figure}

\subsection{Model Architecture}

The overall structure of the SST-GCN model proposed in this paper is shown in Figure 2.(c). The SST-GCN model consists of two main stages: the Spatial Graph Embedding stage and the Temporal Sequence Concatenation stage. In the Spatial Graph Embedding stage, a GCN-based block is used to extract spatial information from the road graph within each time slice. Additionally, a Fully Connected Layer is utilized to extract information from the Static Data. In the Temporal Sequence Concatenation stage, the embedded road information vectors containing spatial information from \( n \) roads at \( k \)-minute intervals, obtained from the Spatial Graph Embedding stage, are passed to an LSTM to extract the temporal information of the roads. Based on the spatiotemporal information obtained through the Spatial Graph Embedding and Temporal Sequence Concatenation stages, SST-GCN uses a Fully Connected Layer and Sigmoid function at the final stage to predict the probability of traffic accident occurrence at a specific node.

\subsection{Assembling GCN and LSTM}
\subsubsection{Spatial Graph Embedding}
In the Spatial Graph Embedding stage, the dynamic feature \( V_{(t-k)} \) is embedded into a vector containing road information through a Graph Convolutional Layer. Within the Spatial Graph Embedding stage, a module consisting of two connected Graph Convolutional Layers is used. The operation of the Graph Convolutional Layer is given by Equation (6), where \( W \) and \( b \) are trainable vectors, and \( V_{(t-k)}^0 = V_{(t-k)} \).

\begin{equation}
        V^{r+1}_{t-k} = LV^{r}_{t-k}W + b
\end{equation}
\begin{center}
    where $r \in [0,1]$
\end{center}

On the other hand, the number of nodes in \( V_{t-k} \) is determined by the structure of the roads extracted based on K-HOP, so the number varies for each sample in the dataset. In other words, the vector \( V_{t-k} \), which has an unfixed shape, is embedded into \( V_{t-k}^2 \), a vector with road spatial information, through two Graph Convolutional Layers, and \( V_{t-k}^2 \) also has an unfixed shape. Since the Temporal Sequence Concatenation stage requires a fixed input shape during the computation process, it is necessary to convert \( V_{t-k}^2 \), embedded from \( V_{t-k} \), into a fixed shape for each sample. To address this, Global Attention Sum Pooling is used after the two Graph Convolutional Layers. The Global Attention Sum Pooling Layer allows us to obtain a fixed vector $X^{dynamic}_{t-k}$ while reflecting the information of the nodes within the graph through the attention mechanism. The operation of the Global Attention Sum Pooling Layer is described in Equations (7) and (8).

\begin{equation}
        \boldsymbol{a} = \text{softmax}(V^{2}_{t-k}\boldsymbol{a})
\end{equation}
\begin{equation}
        X^{dynamic}_{t-k} = \Sigma^{N}_{i=1}a_{i} \cdot v^{2}_{t-k,i}
\end{equation}
\begin{center}
    where $\boldsymbol{a} \in \mathbb{R}$ is a trainable Vector, $v^{2}_{t-k,i} \in V^{2}_{t-k}$
\end{center}

In the Spatial Graph Embedding stage, the Static Data \( S_{t-k} \), composed of static features, is embedded into \( X_{t-k}^{static} \) through two Fully Connected Layers, as shown in Equation (9). The spatial information of the roads, embedded through the two Graph Convolutional Layers and the Global Attention Sum Pooling Layer, is represented as \( X_{t-k}^{dynamic} \). This dynamic vector is concatenated with the static vector \( X_{t-k}^{static} \) as shown in Equation (10), resulting in the vector \( X_{t-k}^{concat} \). Finally, \( X_{t-k}^{concat} \) is transformed into \( X_{t-k}^{Seq} \), a vector encapsulating both dynamic and static information, through two Fully Connected Layers as described in Equation (11). Here, \( W \) and \( b \) are trainable vectors, and \( \sigma \) is the activation function. The activation functions used in the model are detailed in Table 3, and the overall Spatial Graph Embedding stage is illustrated in Figure 2.(a).

\begin{equation}
    X^{static}_{t-k} = \sigma(W\sigma(WS_{t-k}+b)+b)
\end{equation}
\begin{equation}
    X^{concat}_{t-k} = [X^{dynamic}_{t-k}, X^{static}_{t-k}]
\end{equation}
\begin{equation}
    X^{Seq}_{t-k} = \sigma(W\sigma(WX^{concat}_{t-k}+b)+b)
\end{equation}

\subsubsection{Temporal Sequence Concatenation}

The vector \( X_{t-k}^{Seq} \), obtained in the Spatial Graph Embedding stage, encapsulates both the dynamic and static information of the road at time \( t-k \). By extracting \( n \) vectors \( X_{t-k}^{Seq} \) at regular time intervals \( k \), and organizing them sequentially as shown in Equation (12), we aim to consider the temporal correlations between the spatial characteristics of the roads over time.

\begin{equation}
    X^{Seq} = [X^{Seq}_{t-(n-1)k}, ..., X^{Seq}_{t-2k}, X^{Seq}_{t-k}, X^{Seq}_{t}]
\end{equation}

The vector \( X^{Seq} \) obtained through the above process contains both the spatial and temporal information of the road. To more effectively capture the temporal information of \( X^{Seq} \), we use an LSTM. The computational process of the LSTM is given by Equations (13)-(18). Here, \( W_f \), \( W_i \), \( W_C \), \( W_o \), \( b_f \), \( b_i \), \( b_C \), and \( b_o \) are all trainable vectors, \( \sigma \) is the activation function, and \( * \) denotes element-wise multiplication.

\begin{equation}
    i_{q} = \sigma(W_{i}[h_{q-1}, X^{Seq}_{t-(n-q)k}] + b_{i})
\end{equation}

\begin{equation}
    f_{q} = \sigma(W_{f}[h_{q-1}, X^{Seq}_{t-(n-q)k}] + b_{f})
\end{equation}

\begin{equation}
    o_{q} = \sigma(W_{o}[h_{q-1}, X^{Seq}_{t-(n-q)k}] + b_{o})
\end{equation}

\begin{equation}
    \tilde{C}_{q} = tanh(W_{c}[h_{q-1}, X^{Seq}_{t-(n-q)k}] + b_{C})
\end{equation}

\begin{equation}
    C_{q} = f_{q} * C_{q-1} + i_{q} * \tilde{C}_{q}
\end{equation}

\begin{equation}
    h_{q} = o_{q} * tanh(C_{q})
\end{equation}

\begin{center}
    where $q = 1, 2, 3, ..., n$
\end{center}

Through the above process, we can extract the vector \( h_n \), which encapsulates the spatiotemporal information of the road. Finally, using the obtained \( h_n \), we predict the probability \( \hat{y} \) of a traffic accident occurring on a specific road at time \( t \) through two Fully Connected Layers as shown in Equation (19). Here, \( W \) and \( b \) are trainable vectors, and \( \sigma \) is the activation function.

\begin{equation}
    \hat{y} = sigmoid(W\sigma(Wh_{n} + b) + b)
\end{equation}

Consequently, through the Spatial Graph Embedding stage and the Temporal Sequence Concatenation stage, the SST-GCN proposed in this paper can consider not only the spatial information but also the temporal correlations between spatial information. The trained model, using data from \( (n-1)k \) minutes before, at \( k \)-minute intervals, can predict the probability of an accident occurring at time \( t \). The overall Temporal Sequence Concatenation stage is detailed in Figure 2.(b).

\subsection{Training Details}

The model development and training were conducted on Ubuntu 20.04.6 LTS using a Tesla V100-SXM2-32GB GPU. The dataset and model were built using TensorFlow 2.10\cite{21} and the Spektral Library\cite{22}. During the training process, the Adam optimizer (learning rate = 1e-3, beta\_1 = 0.9, beta\_2 = 0.999, epsilon = 1e-7, decay = 2e-6) was used, with a batch size of 32. To prevent the model from overfitting, Early Stopping (monitor = "val\_auc", patience = 10) was employed, and the best-performing model based on validation AUC was selected for testing. The model was trained for a total of 100 epochs. The dataset used for training was split into Train/Validation/Test sets in an 8:1:1 ratio to validate performance, and each set was balanced to have a 1:1 ratio of accident data to non-accident data.

\subsection{Hyper-Parameter Optimization}
\subsubsection{K-HOP, Sequence Number, Time Interval Test}
The factors determining the dataset are K-HOP, the number of sequences \( n \), and the time interval between sequences \( k \). To select the dataset that yields the best performance, a Grid Search was conducted. For each combination of the three factors, the dataset was extracted and trained 10 times for 100 epochs each. The selection of the dataset was based on the average results of these training sessions. The metrics used for dataset selection were Loss (Binary Cross Entropy), F1 Score, Binary Accuracy (threshold = 0.5), and AUC. As a result, the combination of K-HOP = 2, Sequence Number \( n \) = 3, and Time Interval \( k \) = 5 showed superior performance in three out of the six metrics: Precision, Binary Accuracy, and AUC. Detailed results scan be found in Appendix A.

\subsubsection{Adjacency Matrix Filter}
In graph data, the adjacency matrix is crucial as it represents the relationships between nodes, and its preprocessing method significantly impacts the model's performance. Based on the selected parameters of K-HOP = 2, Sequence Number \( n \) = 3, and Time Interval \( k \) = 5 from section 4.4.1, the dataset was extracted to compare the performance of different adjacency matrix preprocessing methods. Experiments were conducted using the Adjacency Matrix \( A_{adjacent} \), which represents node adjacency with 0 and 1, and the Distance Matrix \( A_{distance} \), which uses the Floyd-Warshall algorithm to assign weights based on the distances between nodes. In each matrix, the identity matrix \( I \) was added to consider each node's characteristics during the GCN operations. Subsequently, performance comparison experiments were conducted using the GCN Filter and the Normalized Laplacian Filter on both types of matrices, resulting in four different methods. The matrix \( \bar{A} \) with the GCN Filter applied to matrix \( A \) is given by Equation (20).

\begin{equation}
    \bar{A} = D^{-1/2}AD^{-1/2}
\end{equation}

Similar to section 4.4.1, training was conducted for each adjacency matrix preprocessing method, and the preprocessing method was selected based on the average results of the training sessions. As a result, it was found that applying the Normalized Laplacian Filter to the Distance Matrix \( A_{distance} \), obtained by applying the Floyd-Warshall algorithm to the Adjacency Matrix, resulted in superior model performance. Detailed results can be found in Appendix B.

\subsubsection{Hyper Parameter Optimization Result}

\begin{table}[]
    \centering
    \caption{Details of the Hyper Parameter Search Space in SST-GCN}
    \begin{tabular}{ll}
         \toprule
         Layers & Hyper Parameters \\
         \midrule
         \textbf{GCNConv} & \textbf{units}: [8, 16, 32, 64, 128] \\
         \textbf{FC Layer} & \textbf{units}: [8, 16, 32, 64, 128] \\
         \textbf{LSTM} & \textbf{units}: [8, 16, 32, 64] \\
         \bottomrule
    \end{tabular}
    \label{Details of the Hyper Parameter Search Space in SST-GCN}
\end{table}

\begin{table}[]
    \centering
    \caption{Result of the Hyper Parameter Optimization in SST-GCN}
    \begin{tabular}{ll}
         \toprule
         Layers & Hyper Parameters \\
         \midrule
         \textbf{GCNConv\_1} & \textbf{channels}=64, \textbf{activation}=PReLU \\
         \textbf{GCNConv\_2} & \textbf{channels}=32, \textbf{activation}=PReLU \\
         \textbf{Static\_FC\_1} & \textbf{units}=32, \textbf{activation}=ReLU \\
         \textbf{Static\_FC\_2} & \textbf{units}=16, \textbf{activation}=ReLU \\
         \textbf{Concat\_FC\_1} & \textbf{units}=32, \textbf{activation}=ReLU \\
         \textbf{Concat\_FC\_2} & \textbf{units}=16, \textbf{activation}=ReLU \\
         \textbf{LSTM} & \textbf{units}=8 \\
         \textbf{Output\_FC\_1} & \textbf{units}=8, \textbf{activation}=ReLU \\
         \textbf{Output\_FC\_2} & \textbf{units}=1, \textbf{activation}=ReLU \\
         \bottomrule
    \end{tabular}
    \label{Result of the Hyper Parameter Optimization in SST-GCN}
\end{table}

Based on the results obtained from the experiments in sections 4.4.1 and 4.4.2, Hyper Parameter Optimization was conducted to improve the model's performance. The details of the Hyper Parameter Search Space are shown in Table 2, and the Hyper Parameter Optimization Results can be found in Table 3.

\section{Experiment}
\subsection{Baselines}
To compare the performance of the SST-GCN model, a total of five models were selected. For all models except the Machine Learning Models, the number of epochs was set to 100, and the comparison was based on the average results from 10 repetitions. The details of each comparison model are as follows.

\paragraph{Machine Learning Models}
As comparison models, we used a total of four Machine Learning techniques: Logistic Regression (LR), Decision Tree (DT), Support Vector Machine (SVM), and Random Forest (RF). For Logistic Regression, max\_iter was set to 1000. The construction details for the other models followed the default configurations of the Scikit Learn Library\cite{23}. Since the dataset proposed in this paper has varying shapes for each sample, it cannot be used directly for training these models, as they require a fixed input shape. Additionally, Machine Learning models are not well-suited for learning spatio-temporal information. Therefore, for accident data, only the node information from \( k \) minutes before the accident at the accident location is extracted. For non-accident data, information is randomly extracted from various times and roads to construct the dataset for model evaluation. The Train/Validation/Test sets are constructed in the same manner as described in this paper.

\paragraph{LSTM-CNN}
The model structure and hyperparameters proposed by LSTM-CNN were used as is. The rate for the Dropout Layer was set to 0.2. The dataset features for training the LSTM-CNN were composed of the features proposed in this paper. Additionally, the dataset's timestep was set to 3, resulting in a dataset shape of (3, 38). The optimizer was set to Adam with a learning rate of 1e-3, and the batch size was set to 32.

\paragraph{DST-GCN}
The DST-GCN model can train on datasets with shapes that are not fixed, extracted based on k-hop. Therefore, the dataset constructed in this paper is adapted and used for training and evaluating the DST-GCN model. The detailed preprocessing methods, model structure, and training methods all follow the DST-GCN paper.

\paragraph{MG-TAR}
The MG-TAR model requires a fixed input shape. Therefore, zero padding was applied to the graphs in the dataset constructed in this paper to match the shape of the graph with the maximum number of nodes, ensuring all graphs have the same shape for training. During training, a gradient exploding problem was encountered, so the learning rate for MG-TAR was adjusted to 1e-4. All other details follow the MG-TAR paper.

\subsection{Evaluation Metrics}
For the evaluation of model performance, five metrics were used: Precision, Recall, F1 Score, Binary Accuracy, and AUC. The threshold value for Binary Accuracy was set to 0.5.

\section{Experimental Results}

\begin{table}
    \caption{Performance Comparison for each Model}
    \centering
    \begin{tabular}{l c c c c c}
        \toprule
        Model & Precision & Recall & F1-Score & Binary Accuracy & AUC \\
        \midrule
        LR & 0.5937 & 0.6291 & 0.6109 & 0.5993 & 0.6354 \\
        DT & 0.5666 & 0.5731 & 0.5698 & 0.5673 & 0.5673 \\
        SVM & 0.6042 & 0.6625 & 0.6320 & 0.6142 & 0.6463 \\
        RF & 0.6275 & 0.6872 & 0.6560 & 0.6397 & 0.6950 \\
        LSTM-CNN & 0.5756 & 0.6101 & 0.5843 & 0.5796 & 0.6149 \\
        DST-GCN & 0.7034 & 0.7153 & 0.7082 & 0.7056 & 0.7749 \\
        MG-TAR & 0.7170 & \textbf{0.7546} & 0.7302 & 0.7278 & 0.8070 \\
        \midrule
        \textbf{Proposed} & \textbf{0.8134} & 0.7395 & \textbf{0.7683} & \textbf{0.7837} & \textbf{0.8621} \\
        \bottomrule
    \end{tabular}
    \label{Performance Comparison for each Model}
\end{table}

\begin{figure}
    \centering
    \caption{ROC Curves for each model}
    \includegraphics{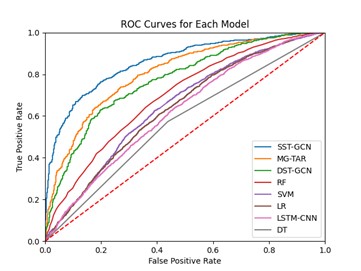}
    \label{ROC Curves for each model}
\end{figure}

The training results of the SST-GCN and comparison models are shown in Table 4, and the ROC Curves for each model are presented in Figure 3. The SST-GCN model achieved the highest performance in four out of five metrics: Precision, F1 Score, Binary Accuracy, and AUC. Experimentally, the SST-GCN identified K-HOP=2, Sequence Number n=3, and Time Interval k=5 as the most suitable parameters for predicting road risk. It also demonstrated that applying the normalized Laplacian filter yields superior performance. This suggests that meaningful interactions occur between roads within a K-HOP=2 range rather than over a wide area, and that reflecting road conditions within short intervals (Sequence Number n=3, Time Interval k=5) is most effective for predicting risk.

The SST-GCN model outperformed all four Machine Learning models in every metric. This can be attributed to its design, which converts road information into a sequential graph form and utilizes graph convolution layers and LSTM, enabling the model to effectively capture the graph-based characteristics of roads.

The LSTM-CNN model applied LSTM to reflect temporal characteristics but only used information from a single road for predictions, failing to capture the complex interactions between roads, resulting in performance similar to the Machine Learning models.

The DST-GCN model effectively captured the graph characteristics of roads by performing graph convolution operations on the graph information consisting of the central road and adjacent roads extracted based on K-HOP. However, it predicted accident risk using only the embedding vector of the central node after graph convolution, without utilizing RNN layers, leading to lower performance compared to SST-GCN. The SST-GCN model, on the other hand, compressed information more effectively using Global Attention Sum Pooling on the embedding vectors of all nodes, rather than only the central node, leading to superior performance over DST-GCN.

Lastly, the MG-TAR model showed better performance than other comparison models by effectively reflecting the spatiotemporal characteristics of roads through graph convolution operations and a multi-head attention module. However, the process of dividing graph features, computing new features for each node through different graph convolutional layers, and extracting information again through graph convolution layers increased the model's complexity, making it challenging to train.

\section{Conclusion}
This paper proposes the Sequential based Spatio-Temporal Graph Convolutional Networks for Minute-level and Road-level Traffic Accident Prediction (SST-GCN). To enhance the model's performance, we compared model performance across datasets extracted based on different K-HOPs, sequence numbers, and time intervals between sequences, selecting the most suitable dataset. Additionally, we compared the extraction and preprocessing methods of adjacency matrices, which represent the relationships between nodes in the Graph Convolutional Network, to choose the most effective method for improving model performance. The SST-GCN model was designed to capture the unique dynamic features and common static features of each road at specific times, and to capture the temporal correlations between the spatial features of roads. As a result, the SST-GCN model outperformed comparison models in four metrics, demonstrating its effectiveness in traffic accident prediction. Consequently, the SST-GCN model proposed in this paper is expected to be practically used in various intelligent transportation systems for traffic accident prediction and prevention at the Minute-Level and Road-Level.

\section{Acknowledgement}

%Bibliography
\bibliographystyle{unsrt}  
\bibliography{references}  

\clearpage
\begin{appendices}
\section{Performance Comparison according to K-HOP, Sequence Number, Time Interval}
\centering
\begin{tabular}{c|c|c|c|c|c|c}
    \toprule
        KHOP/SeqNum/Interval & Loss & Precision & Recall & F1-Score & Binary Accuracy & AUC \\
    \midrule
        1/2/5	& 0.5082	& 0.7870	& 0.7010	& 0.7315	& 0.7546	& 0.8397 \\
        1/2/10	& 0.4979	& 0.7703	& 0.7613	& 0.7576	& 0.7656	& 0.8467 \\
        1/2/15	& 0.4772	& 0.7833	& 0.7442	& 0.7584	& 0.7682	& 0.8546 \\
        1/3/5	& 0.4875	& 0.7539	& 0.7864	& 0.7594	& 0.7633	& 0.8500 \\
        1/3/10	& 0.5164	& 0.7594	& 0.7305	& 0.7372	& 0.7488	& 0.8335 \\
        1/3/15	& 0.5105	& 0.7723	& 0.7402	& 0.7489	& 0.7591	& 0.8390 \\
        1/4/5	& 0.5001	& 0.7707	& 0.7557	& 0.7535	& 0.7641	& 0.8425 \\
        1/4/10	& 0.5031	& 0.7731	& 0.7402	& 0.7503	& 0.7604	& 0.8397 \\
        1/4/15	& 0.5073	& 0.7552	& 0.7656	& 0.7516	& 0.7572	& 0.8419 \\
        2/2/5	& 0.4976	& 0.7812	& 0.7400	& 0.7508	& 0.7654	& 0.8479 \\
        2/2/10	& 0.4854	& 0.7665	& 0.7858	& 0.7681	& 0.7714	& 0.8547 \\
        2/2/15	& 0.4848	& 0.7859	& 0.7475	& 0.7597	& 0.7715	& 0.8537 \\
        \textbf{\underline{2/3/5}}	& 0.4758	& \textbf{\underline{0.8134}}	& 0.7395	& 0.7683	& \textbf{\underline{0.7837}}	& \textbf{\underline{0.8621}} \\
        2/3/10	& 0.4853	& 0.7896	& 0.7570	& 0.7633	& 0.7766	& 0.8568 \\
        2/3/15	& 0.4753	& 0.7724	& 0.7843	& 0.7718	& 0.7772	& 0.8588 \\
        2/4/5	& 0.4801	& 0.7665	& 0.7748	& 0.7620	& 0.7681	& 0.8564 \\
        2/4/10	& 0.5085	& 0.7776	& 0.7347	& 0.7465	& 0.7595	& 0.8450 \\
        2/4/15	& 0.4991	& 0.7857	& 0.7295	& 0.7476	& 0.7628	& 0.8490 \\
        3/2/5	& \textbf{\underline{0.4741}}	& 0.7853	& 0.7741	& 0.7733	& 0.7819	& 0.8601 \\
        3/2/10	& 0.4850	& 0.7765	& 0.7729	& 0.7670	& 0.7738	& 0.8554 \\
        3/2/15	& 0.4993	& 0.7728	& 0.7755	& 0.7675	& 0.7724	& 0.8447 \\
        3/3/5	& 0.4966	& 0.7628	& 0.7937	& 0.7686	& 0.7720	& 0.8518 \\
        3/3/10	& 0.5037	& 0.7901	& 0.7607	& 0.7661	& 0.7770	& 0.8453 \\
        3/3/15	& 0.4781	& 0.7716	& 0.7781	& 0.7690	& 0.7733	& 0.8569 \\
        3/4/5	& 0.5098	& 0.7779	& 0.7464	& 0.7565	& 0.7662	& 0.8505 \\
        3/4/10	& 0.5115	& 0.7820	& 0.7603	& 0.7675	& 0.7732	& 0.8494 \\
        3/4/15	& 0.5085	& 0.7732	& 0.7421	& 0.7486	& 0.7603	& 0.8392 \\
        4/2/5	& 0.5131	& 0.7626	& 0.7468	& 0.7513	& 0.7539	& 0.8384 \\
        4/2/10	& 0.4946	& 0.7900	& 0.7376	& 0.7586	& 0.7667	& 0.8546 \\
        4/2/15	& 0.5266	& 0.7470	& 0.8005	& 0.7733	& 0.7605	& 0.8438 \\
        4/3/5	& 0.4914	& 0.7700	& \textbf{\underline{0.8125}}	& \textbf{\underline{0.7795}}	& 0.7809	& 0.8561 \\
        4/3/10	& 0.5032	& 0.8013	& 0.7024	& 0.7349	& 0.7596	& 0.8415 \\
        4/3/15	& 0.5129	& 0.7759	& 0.7618	& 0.7626	& 0.7671	& 0.8478 \\
        4/4/5	& 0.5125	& 0.7840	& 0.7612	& 0.7679	& 0.7746	& 0.8395 \\
        4/4/10	& 0.5608	& 0.7378	& 0.7595	& 0.7360	& 0.7378	& 0.8128 \\
        4/4/15	& 0.5440	& 0.6668	& 0.6769	& 0.6684	& 0.7201	& 0.7924 \\
    \bottomrule
\end{tabular}

\section{Performance Comparison according to Adjacency Matrix Extraction Methods and Filters}
\centering
\resizebox{\textwidth}{!}{
\begin{tabular}{l|c|c|c|c|c|c}
    \toprule
        Preprocessing & Loss & Precision & Recall & F1-Score & Binary Accuracy & AUC \\
    \midrule
        Adjacent Matrix + GCN Filter	                & 0.4879 & 0.7837 & 0.7370 & 0.7508 & 0.7656 & 0.8492 \\
        Adjacent Matrix + Normalized Laplacian Filter	& 0.5051 & 0.7555 & \textbf{\underline{0.7566}} & 0.7469 & 0.7542 & 0.8393 \\
        Distance Matrix + GCN Filter	                & 0.515	 & 0.7631 & 0.7326 & 0.7378 & 0.7514 & 0.8319 \\
        Distance Matrix + Normalized Laplacian Filter	& \textbf{\underline{0.4758}} & \textbf{\underline{0.8134}} & 0.7395 & \textbf{\underline{0.7683}} & \textbf{\underline{0.7837}} & \textbf{\underline{0.8621}} \\

    \bottomrule
\end{tabular}
}
\end{appendices}

\end{document}